\def\year{2019}\relax
\documentclass[letterpaper]{article} 
\usepackage{aaai19}  
\usepackage{times}  
\usepackage{helvet}  
\usepackage{courier}  
\usepackage{url}  
\usepackage{graphicx}  
\usepackage{booktabs}       
\usepackage{amsfonts}       
\usepackage{nicefrac}       
\usepackage{microtype}      
\usepackage{amsmath}
\usepackage{subcaption}
\usepackage{algorithm}
\usepackage{algorithmicx}
\usepackage{algpseudocode}
\usepackage{multirow}
\usepackage{bbold}
\usepackage{mathtools}
\DeclareMathSizes{10}{7.5}{4}{4}
\begin{document}
\medmuskip=0mu
\thinmuskip=0mu
\thickmuskip=0mu
%
\title{SCNN: A General Distribution based Statistical Convolutional Neural Network with Application to Video Object Detection}
\author{
    Tianchen Wang\textsuperscript{\rm 1}, Jinjun Xiong\textsuperscript{\rm 2}, Xiaowei Xu\textsuperscript{\rm 3}, Yiyu Shi\textsuperscript{\rm 4}\\
    \textsuperscript{\rm 1,3,4}Department of Computer Science and Engineering, University of Notre Dame\\
    \textsuperscript{\rm 2}IBM Thomas J. Watson Research Center \\
    \{\textsuperscript{1}twang9, \textsuperscript{3}xxu8, \textsuperscript{4}yshi4\}@nd.edu, \textsuperscript{2}jinjun@us.ibm.com\\
}
\maketitle

\begin{abstract}
Various convolutional neural networks (CNNs) were developed recently that achieved accuracy comparable with that of human beings in computer vision tasks such as image recognition, object detection and tracking, etc. Most of these networks, however, process one single frame of image at a time, and may not fully utilize the temporal and contextual correlation typically present in multiple channels of the same image or adjacent frames from a video, thus limiting the achievable throughput. This limitation stems from the fact that existing CNNs operate on deterministic numbers. In this paper, we propose a novel statistical convolutional neural network (SCNN), which extends existing CNN architectures but operates directly on correlated distributions rather than deterministic numbers. By introducing a parameterized canonical model to model correlated data and defining corresponding operations as required for CNN training and inference, we show that SCNN can process multiple frames of correlated images effectively, hence achieving significant speedup over existing CNN models. We use a CNN based video object detection as an example to illustrate the usefulness of the proposed SCNN as a general network model. Experimental results show that even a non-optimized implementation of SCNN can still achieve 178\% speedup over existing CNNs with slight accuracy degradation. 
\end{abstract}
\section{Introduction}
\noindent
With strong feature extraction capabilities from deep convolutional neural networks (CNNs) and many optimized
implementations \cite{xu2017edge} of the associated deep learning frameworks, the performance of various computer vision tasks has been drastically improved. 
For example in image recognition and object detection, deep CNN architectures as ResNet \cite{he2016deep}, DenseNet \cite{huang2017densely} and frameworks as YOLO \cite{redmon2017yolo9000}, faster R-CNN \cite{ren2015faster} all outperform then state-of-the-art by an impressive margin at the time of their publication.

However, a mainstay for any CNN based framework is that it operates on  deterministic weights and inputs \cite{xu2018scaling,xu2018quantization}. 
These frameworks process one image at a time during both training and inference. This is obviously not ideal as it largely ignores both temporal and contextual correlation
existing among channels and
adjacent frames.
To break this mainstay, in this paper we propose to explicitly
model such correlation by 
extracting parameterized canonical distributions from correlated inputs (such as adjacent frames in a video), and design a statistical convolutional neural network (SCNN) to propagate these correlated distributions directly.
Our SCNN can be easily integrated into existing CNN architectures by replacing their deterministic operations with our statistical
counterparts operating on parameterized canonical distributions.  
Then with little modification to the existing gradient descent based scheme, our SCNN can be trained using the same forward and back propagation procedures. 

More specifically, we first build a linear parameterized canonical form via independent component analysis (ICA) to represent the statistical distribution of each input component to capture its temporal and contextual correlation. 
We then define all the required CNN operations (such as convolution, ReLU, batch normalization etc.) in terms of the parameterized canonical form, including their various partial gradients for backward propagation, thus enabling us to integrate SCNN with any existing CNN implementation frameworks easily. 
To show the effectiveness of the proposed SCNN, we further apply it to video object detection task and propose a new objective function that improves SCNN based training. 
Even though many great successes have been achieved in objection detection for static images \cite{ren2015faster,redmon2017yolo9000,lin2018focal,liu2016ssd}, the performance of video object detection still has a large room for improvement, especially for its real-time throughput.
Since its introduction in ImageNet competition, multiple solutions have been proposed \cite{kang2017t,han2016seq}, most of which solve the problem by extending static image object detection methods to consider adjacent frames for temporal information.
Their efficiency is, however, not satisfactory for online detection, and for training, it would take several days to generate video tubelets \cite{kang2017t}. 
A recent research \cite{zhu2017flow} proposed a flow-guided feature aggregation where it considers adjacent frames at feature level rather than at the box level, but it requires repeated sampling and complex modeling.
It is still desirable to have a more direct and effective way of modeling correlated adjacent frames.

Our main contributions in this paper are as follows.
1) We propose a novel statistical convolutional neural network that can act as a backbone alternative to any existing CNN architectures and operates directly on distributions rather than deterministic numbers, 
2) We use a parameterized canonical model to capture correlated input data for CNN and reformulate popular CNN layers to adapt their forward
and backward computation for parameterized canonical models.
3) We adopt video object detection as an examplar application and introduce a new objective function for better training of SCNN. 
4) We conduct experiments on an industrial UAV object detection dataset and show that SCNN backbone can achieve up to $178\%$ speedup over conventional counterpart with slight accuracy degradation.

\section{Review of ICA}
ICA is a well-known technique in signal processing to separate a multivariate signal into a set of additive random subcomponents that are statistically independent from each other. 
The random subcomponents are typically modeled as non-Gaussian distributions. 
In some cases, a priori knowledge of the probability distributions of these random subcomponents can be also incorporated into ICA. 
The random subcomponents are also called the basis of the corresponding multivariate signal.
We denote an n-dimensional multivariate signal as a random vector $D = (D_1, D_2, \dots, D_n)^T$. 
The random subcomponents are denoted as a random vector $X = (X_1, X_2, \dots, X_m)^T$. 
For a given set of $N$ samples (realizations) of the multivariate signal's random vector $D$, each component $D_i$ of the $N$ samples can be treated as being generated by a sum of some realization of the $m$ independent random subcomponents, which is given by
\begin{equation}
D_i = a_{i,1} X_1 + a_{i,2} X_2 + \dots + a_{i,k} X_k + \dots + a_{i,m} X_m, ~k\in\{1, m\}
\end{equation}
\noindent where $a_{i,k}$ is the mixing weight of the corresponding random subcomponent $X_k$. 
We can put them compactly in a matrix form as follows
\begin{equation}
    D = A X
\end{equation}
where $A$ is the mixing matrix.
The goal of ICA is to estimate both the mixing matrix $A$ and the corresponding realization of the random subcomponent $X$ (i.e., $X = WD$). 
The realization of the basis $X$ can be obtained either by inverting $A$ directly (i.e., $W = A^{-1}$) or through the pseudo inverse of $A$.

The ICA has also been extended to consider the case where a zero-mean uncorrelated Gaussian noise $R_i$ is added. 
Without loss of generality, we can normalize all basis (random subcomponents) to have a zero mean and standard deviation of 1. 
In other words, we have
\begin{equation}
D_i = a_{i,0} + a_{i,1} X_1 + a_{i,2} X_2 + \dots + a_{i,m} X_m + a_{i,r} R_i \label{eq:ICA}
\end{equation}
where $a_{i,0}$ is the mean value for $D_i$, and $a_{i,r}$ is the weighting of the modeled uncorrelated Gaussian noise term.

\begin{figure}[!t]
  \centering
  \includegraphics[width=6cm]{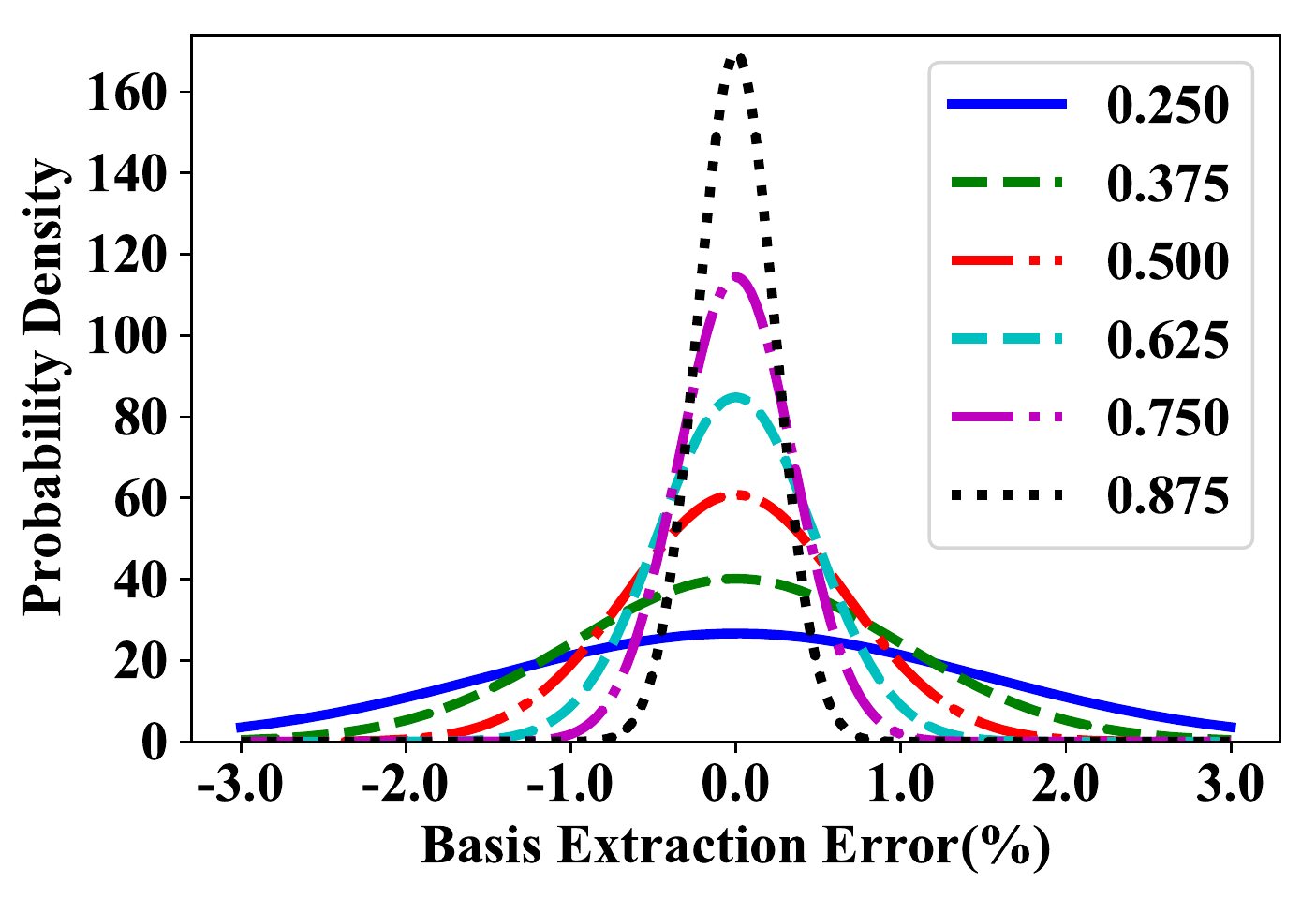}
    \caption{ICA modeling error (\%) with different $m/N$.}
  \label{fig:error_hist}
\end{figure}
\begin{figure*}[!t]
  \centering
  \includegraphics[width=13cm]{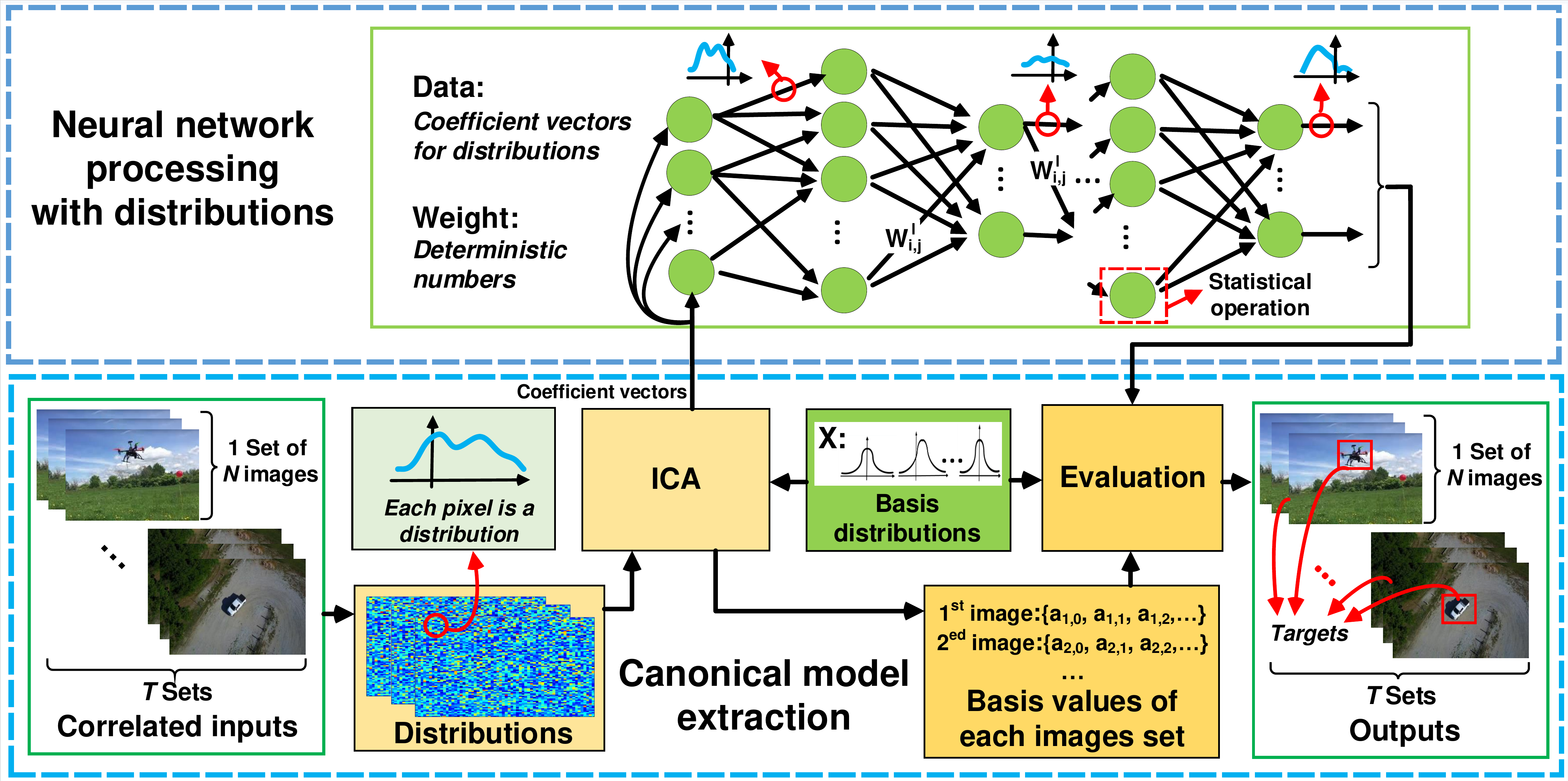}
    \caption{Overall structure of SCNN (illustrating an object detection task in a video).}
  \label{fig:scnn}
\end{figure*}

\section{Statistical Convolutional Neural Network}
\label{sec:scnn}
\subsection{Correlated Inputs Modeling}
Many existing applications with CNN models have inputs that exhibit strong temporal and contextual (spatial) correlations, such as multiple adjacent frames in a video stream. 
Therefore, we can model these inputs as a multivariate signal. 
For a given set of $N$ samples (realizations) of the inputs, such as multiple correlated frames of a video snippet, we can represent each component of the inputs via ICA as a linear additive model of a set of $m$ independent random subcomponents as shown in Equation~(\ref{eq:ICA}).
In the rest of the paper, we define $N$ as extraction span and $m$ as the basis dimension. 
Moreover, because the random subcomponents $X_1$, $\dots$, $X_m$ are shared among all input components, we can use the above model to compactly capture both the temporal and contextual correlations. 
We call such a model as a linear parameterized canonical model, and the weights such as $a_{i,k}$ as parameters.

To demonstrate the accuracy of ICA to model correlated frames in a video, we extract the distributions from a few small video snippets in our experiment dataset \cite{uavdata} and depict the error distribution between the original data and the unmixing result in Figure~\ref{fig:error_hist}. From the figure we can see that increasing the ratio of $m/N$ in general reduces the error, and the error is mostly bounded by $3\%$.

In the context of CNN, this type of modeling of correlated inputs raises a number of interesting questions. 
(1) For a given trained CNN network with model parameters, how do we carry out the inference for such a parameterized canonical model? 
(2) How to train such a CNN network with each input being represented as a parameterized canonical form?
We will provide answers to address the above two questions in the rest of the paper. 

Because of the way we model the inputs as a parameterized statistical distribution, we call our network as SCNN.
With parameterized canonical model, the overall structure of our SCNN is illustrated in Figure~\ref{fig:scnn}. With details discussed in the following sections, the input video stream is divided into snippets, each containing a number of correlated adjacent frames. Those images in the same snippet will be modeled by one image of the original image size, but each pixel of which is taking as a canonical form. These canonical forms are forward and back propagated through CNN for both training and inference. At the output of the network, all the canonical forms are converted to the corresponding scalar values by plugging the estimated realizations
of random subcomponents ($X = WD$) of each input video snippet. With that, 
we obtain a feature map with scalar values, hence the conventional objective function evaluation can be carried out.

\subsection{Forward Propagation in SCNN}
In a typical CNN network, there are a number of commonly used layers, such as fully connected layer (FC), convolutional layer (CN), ReLU layer, max-pooling layer, batch normalization layer. 
We will provide the corresponding implementation details in SCNN for these commonly used layers in the following. Again we would like to emphasize  that the discussion here does not restrict to any particular CNN architecture. In our experiments we will demonstrate our implementation on various CNN architectures.

Before we delve into the details of each layer implementation, we note that there are two core operations
for these layers: 
(1) a weighted \textit{sum} of a set of input numbers (which is used frequently for both FC and CN layers), and 
(2) a \textit{max} of a set of input numbers (which is frequently used for both ReLU and max-pooling layers). 
In SCNN, the input numbers to the above two operations are no longer deterministic numbers, but parameterized statistical distributions. 
We discuss how we provide solutions to these two core operations first. 
Please note that, some of the discussion related to the sum and max operations has been covered in prior literature in the area of statistical timing analysis \cite{xiong2008incremental,cheng2009non,visweswariah2006first,singh2006statistical}. 
We obtain a lot of inspiration from their work. 
We only repeat essential points in this paper for completeness, but refer interested readers to those references for greater details and proofs.

\subsubsection{Sum operation}
For two inputs $D_i$ and $D_j$, their sum can be represented as
\begin{equation}
\begin{split}
D_{\text{sum}} &= D_i  + D_j \\ 
&= (a_{i,0} + a_{j,0}) + (a_{i,1} + a_{j,1}) X_1 + (a_{i,2} + a_{j,2}) X_2 + \dots \\
&+ (a_{i,m} + a_{j,m}) X_m + a_{i,r}R_i + a_{j,r}R_j.
\end{split}
\end{equation}

As we can see, the \textit{sum} operation as defined above will give us back a similar parameterized canonical form. 
This is important as it allows us to carry out similar operations repeatedly across layers. 
Because of this, for multiple inputs, similar sum operations can be applied easily.
Most interestingly, the computation involves only those parameters, but not random subcomponents.

\subsubsection{Max operation}
The \textit{max} operation is a bit more involved. 
We start with the most common scenario where the distribution of random subcomponents $X_1$, $\dots$, $X_m$ are modeled as a Gaussian distribution. 
In this case, for two inputs $D_i$ and $D_j$, their max can be represented as
\begin{equation}
D_{\text{max}} = max(D_i, D_j) = a_{\text{max},0} + \sum_{k=1}^{m}a_{\text{max},k} X_k +  a_{\text{max},r} R_{\text{max}}    
\end{equation}
where $a_{max,0}$ and $a_{max,r}$ are obtained by matching the first and $2^{nd}$ moments of the above equation on both sides;
$a_{max,k}$ are obtained via
the tightness probabilities as introduced in \cite{visweswariah2006first} to represent the probability that one distribution is larger than (or dominates) the other given by
\begin{equation}
t_i = \int_{-\infty}^{\infty}\frac{1}{\sigma_{D_i}}\phi(\frac{x-a_{i, 0}}{\sigma_{D_i}})\Phi(\frac{(\frac{x-a_{j, 0}}{\sigma_{D_j}})-\rho(\frac{x-a_{i, 0}}{\sigma_{D_i}})}{\sqrt{1-\rho^2}})dx = \Phi(\beta)
\end{equation}
where $\theta$, $\beta$, $\phi$ and $\Phi$ are defined as 
\begin{equation}
\begin{split}
&\theta=\sqrt{\sigma_{D_i}^2+\sigma_{D_j}^2-2\sigma_{D_i}\sigma_{D_j}},\hspace{2em} \beta=(\frac{a_{i, 0}-a_{j, 0}}{\theta}),\\
&\phi(x)=\frac{1}{\sqrt{2\pi}}exp(-\frac{x^2}{2}), \hspace{2em}\Phi(y)=\int_{-\infty}^y\phi(x)dx.
\end{split}
\end{equation}
Therefore, the mean $a_{\text{max}, 0}$ and variance $\sigma_{D_{\text{max}}}^2$ of $D_{\text{max}}$ can be expressed analytically as
\begin{equation}
\begin{split}
a_{\text{max}, 0} &= a_{i, 0} \Phi(\beta) + a_{j, 0} \Phi(-\beta) + \theta \phi(\beta), \\
\sigma_{D_{\text{max}}}^2 &= (\sigma_{D_{i}}^2+a_{i, 0}^2)\Phi(\beta)+(\sigma_{D_{j}}^2+a_{j, 0}^2)\Phi(-\beta) \\ 
&+(a_{i, 0} + a_{j, 0})\theta\phi(\beta)-a_{\text{max}, 0}^2.
\end{split}
\end{equation}
And the corresponding canonical form of $D_{\text{max}}$ is
\begin{equation}
\begin{split}
D_{\text{max}} &= a_{\text{max}, 0} + \sum_{k=1}^m a_{\text{max}, k} X_k +a_{\text{max}, r} R_{\text{max}} \\
\text{where}\hspace{1em}a_{\text{max}, k} &= \Phi(\beta) a_{i, k} + \Phi(-\beta) a_{j, k}, ~~ k=\{1, m\}, \\
a_{\text{max}, r} &= (\sigma_{D_{\text{max}}}^2-\sum_{k=1}^m a_{\text{max}, k}^2)^{1/2}.
\end{split}
\end{equation}
It is noted that $\sigma_{D_{\text{max}}}^2-\sum_{k=1}^m a_{\text{max}, k}^2$ is proved to be always non-negative by \cite{sinha2005statistical}.
For simplicity, we use 
$\Phi$ and $\phi$ to represent $\Phi(\beta)$ and $\phi(\beta)$, and 
\textit{sum} and \textit{max} to represent the statistical operations between distributions, and the notations will be used wherever there is no ambiguity.

Same as the \textit{sum} operation, the \textit{max} operation as defined above will give us back a similar parameterized canonical form. 
This is important as it allows us to carry out similar max operations repeatedly across layers. 
Because of this, for multiple inputs, we can repeatedly apply the two input max operations and obtain the final multi-input max result, i.e., 
\begin{equation}
\textit{max} (D_1, D_2, \dots, D_p) = \textit{max}(D_1, \textit{max}(D_2, \textit{max} (D_3,\dots.)))).    
\end{equation}

In the same spirit, more sophisticated approaches to handle the \textit{max} operation of multiple inputs and non-Gaussian
distributions have been discussed in references such as \cite{xiong2006criticality,mogal2007clustering}. 
In the interests of space, we'll not repeat it here.
\subsubsection{FC \& CN layers}
Key to the two layers' operation is the computation of a weighted \textit{sum}. 
When the inputs are parameterized canonical form, we can decompose the weighted sum in two logic steps: 
(1) for each input, we scale the input by the weight and obtain a similar canonical form; and 
(2) for the remaining sum operation, it is carried out same as the sum of a set of canonical forms.

For FC, given an input with $p$ distributions $D_i^l (i\in\{1, p\})$ at layer $l$, the forward operation computes the $j$th output distributions $D_j^{l+1} (j\in\{1, q\})$ with weight $w$ as:
\begin{equation}
\begin{split}
    D_{j}^{l+1} &= a_{j, 0}^{l+1}+\sum_{k=1}^m a_{j, k}^{l+1} X_k + a_{j, r}^{l+1}R = \sum_{i=1}^{p}w_{i, j}D^{l}_{i} \\
    &=\sum\limits_{i=1}^{p}w_{i, j}a_{i, 0}^{l}+ \sum\limits_{i=1}^{p}\sum\limits_{k=1}^{m}w_{i, j}a_{i, k}^{l} X_k +\sqrt{\sum\limits_{i=1}^{p}(w_{i, j}  a_{i, r}^{l})^2} R.
\end{split}
\end{equation}

For CN, an input distribution tensor would be given as $D^{l}\in\mathbb{R}^{m+2, h, w}$ ($h\times w$ distributions at layer $l$) and a convolution filter $W\in\mathbb{R}^{x', y'}$ ($x'$, $y'$ denote the convolution kernel mask size) for the next layer,
we have the forward propagation for SCNN CN at position $x, y$ expressed as
\begin{equation}
\begin{split}
    D_{x, y}^{l+1} &= a_{x, y, 0}^{l+1}+\sum_{k=1}^m a_{x, y, i}^{l+1} X_i + a_{x, y, r}^{l+1}R
    = D_{x, y}^{l}\ast W_{x, y}
    \\
    &=\sum\limits_{x'} \sum\limits_{y'} W_{x', y'} a_{x-x', y-y', 0}^{l}+
    \sum\limits_{x'} \sum\limits_{y'}\sum\limits_{k=1}^{m} W_{x', y'} a_{x-x', y-y', k}^{l} X_k \\
    &+\sqrt{\sum\limits_{x'} \sum\limits_{y'}(W_{x', y'}a_{x-x', y-y', r}^l)^2} R
\end{split}
\end{equation}
where $\ast$ denotes the convolution operation.

\subsubsection{ReLU and max-pooling layers}
Since the key operation in both ReLU and max-pooling layers is the \textit{max} operation, we can easily extend the \textit{max} operation as discussed above to handle the canonical inputs. 
In the case of ReLU, the reference point is not necessary to be zero, and it can be defined as a distribution. But in our current implementation, we still choose a constant reference for ReLU.

In max-pooling layer, new distributions are generated from the previous layer distributions under sliding masks with \textit{max} operation.
Given an input distribution tensor $D^l\in\mathbb{R}^{m+2, h, w}$, and a \textit{max} pooling filter $K\in\mathbb{R}^{h_f, w_f}$, the problem is to obtain an output distribution tensor $D^{l+1}$ of \textit{max} distribution from partitioned subtensors.
Therefore the forward propagation with stride $s$ and without padding can be expressed as
\begin{equation}
D_{x, y}^{l+1} = \underset{(x, y)\in[s\times x, s\times x+h_f]\times[s\times y, s\times y+w_f]}{\textit{max}}D_{x, y}^{l}.
\end{equation}
In traditional max-pooling layer, the locations of maximum values at the current layer under kernel masks are stored for back propagation.
During SCNN max-pooling implementation, we store the tightness probabilities between the corresponding distributions during forward propagation, which indicate the contributions of the distributions at the current layer to the ones at the next layer.

\subsubsection{Batch normalization layer}
In SCNN, we do not follow the traditional batch normalization layer \cite{ioffe2015batch} definition. 
Instead, we define the operation as follows in consideration of the canonical inputs:
given an input distribution $D_i^l$ with basis sensitivity $a_{i, k}$ and variance $\sigma_{D_i^l}^2$, the normalization output distribution $D_{i}^{l+1}$ is expressed as
\begin{equation}
\begin{split}
D_i^{l+1} &= a_{i,0}^{l+1} + \sum_{k=1}^{m}a_{i,k}^{l+1} X_k + a_r^{l+1} R_i\\
&=a_{i, 0}^{l} + \sum_{k=1}^{m}(\gamma \frac{a_{i, k}^{l}-\frac{1}{n} \sum_{j=1}^{m}a_{i, j}^{l}}{\sqrt{\sigma_{D_i^l}^2+\epsilon}} + \beta) + a_r^{l} R_i
\end{split}
\end{equation}
where $\gamma$, $\beta$ are the learning coefficients.
Instead of evaluating the values on mini-batch, we perform normalization on each input distribution.
Note that $a_{i, 0}$ and $a_{i, r}$ are not involved in normalization.

\subsubsection{Inference at the output layer}
After we carry out the parameterized forms through the various layers in the CNN network as discussed above, we arrive at the end of the network where we need to decide the output. 
Here we resort to a simple approach, i.e., we convert the canonical forms to their corresponding scalar values by plugging the estimated realizations of random subcomponents $X = W D$. 
With that, we obtain the output layer with scalar values, hence conventional inference at the last output layer can be carried out.

\subsection{Back Propagation in SCNN}
The back propagation is key to the training of the SCNN by computing various gradients of the cost function with respect to network parameters, which in term depends on computing the partial derivatives of various operation outputs with respect to their inputs.

\subsubsection{Partial derivative for sum}
Given two distributions $D_i$, $D_j$ along with two weights $w_i$, $w_j$ ,and the sum $D_{\text{sum}} = w_i D_i + w_j D_j$, the partial derivative of $D_{\text{sum}}$ w.r.t. the sensitivities in $D_i$ is expressed as
\begin{equation}
    \frac{\partial D_{\text{sum}}}{\partial a_{i, k}} = w_i, \hspace{2em} \frac{\partial D_{\text{sum}}}{\partial a_{i, r}} = \frac{w_i}{a_{\text{sum}, r}} 
\end{equation}
where $k\in\{0, m\}$.
Then with the help of gradient of \textit{sum} operation, the derivatives of FC and CN in SCNN are obtained accordingly.
Given the gradient of $D_{j}^{l+1}$ at layer $l+1$ as $\delta_{j}^{l+1}$ ($j\in\{1, q\}$), the gradients of each sensitivities in distribution $D_{i}^{l}$ ($i\in\{1, p\}$) at layer $l$ are shown as
\begin{equation}
\begin{split}
\delta_{i, k}^{l} = \sum\limits_{j=1}^{q} \delta_{j}^{l+1}\frac{\partial D_j^{l+1}}{\partial a_{i, k}^{l}} &= \sum\limits_{j=1}^{q} \delta_{j}^{l+1} w_{i, j}, \hspace{2em} k\in\{0, m\},  \\
\delta_{i, r}^{l} = \sum\limits_{j=1}^{q} \delta_{j}^{l+1}\frac{\partial D_j^{l+1}}{\partial a_{i, r}^{l}} &= \sum\limits_{j=1}^{q} \delta_{j}^{l+1}\frac{w_{i, j}}{a^{l+1}_{n,k}}. 
\end{split}
\end{equation}
The partial derivatives of total cost $L$ w.r.t. corresponding weight $w_{i, j}$ in FC goes to
\begin{equation}
\begin{split}
\frac{\partial L}{\partial w_{i, j}} = \delta_{j}^{l+1} \frac{\partial D_j^{l+1}}{\partial w_{i, j}}=\delta_{j}^{l+1}(\sum\limits_{k=0}^{m} a^{l}_{i, k}+\frac{a^{l}_{i, r}}{a^{l+1}_{j, r}}).
\end{split}
\end{equation}

The derivative of SCNN CN layer follows the same path.
Given the gradient of $D^{l+1}$ w.r.t. total cost $L$ as $\delta^{l+1}$, the gradients of each sensitivities in distribution $D_{x, y}^{l}$ at location $x, y$ as $\delta_{x, y}^{l}$ are shown as
\begin{equation}
\begin{split}
\delta_{x, y, k}^l &= \frac{\partial L}{\partial a_{x, y, k}^l} = \delta_{x, y, k}^{l+1} \ast W_{-x, -y}^{l+1},  \hspace{2em} k\in\{0, m\}, \\
\delta_{x, y, r}^l &= \frac{\partial L}{\partial a_{x, y, r}^l} = \frac{\delta_{x, y, r}^{l+1}a_{x, y, r}^l}{a_{x, y, r}^{l+1}} \ast (W_{-x, -y}^{l+1})^2.
\end{split}
\end{equation}
The gradient of convolution weight is derived as
\begin{equation}
\frac{\partial L}{\partial W_{x, y}^{l+1}} = \sum\limits_{k=0}^{m}\delta_{x, y, k}^{l+1} \ast a_{-x, -y, k}^{l} + \frac{\delta_{x, y, r}^{l+1} W_{x, y}^{l+1}}{a_{x, y, r}^{l+1}} \ast (a_{-x, -y, r}^{l})^2.
\end{equation}
\subsubsection{Partial derivative for max}
The derivative of \textit{max} in distributions is mainly involved in the back propagation of SCNN ReLU and Max-pooling layer.
Given two distributions $D_i$ and $D_j$ with $D_{\text{max}}=\textit{max}(D_i, D_j)$, the gradient of mean and variance of $D_{\text{max}}$ with respect to $D_i$ is derived by \cite{xiong2008incremental}.
We follow the similar routine and derive the gradients of each sensitivities of $D_{\text{max}}$ with respect to the ones of $D_{i}$.
For $p\in\{1, m\}$, we first compute the gradient of $\theta$, $\phi$, $\Phi$ with respect to $a_{i, 0}$, $a_{i, p}$ and $a_{i, r}$.
Then the gradients of $a_{\text{max}, q}$ ($q\in\{1, m\}$), $a_{\text{max}, 0}$ and $a_{\text{max}, r}$ with respect to $a_{i, 0}$, $a_{i, p}$ and $a_{i, r}$ are obtained as
\begin{equation}
\begin{split}
\frac{\partial a_{\text{max}, 0}}{\partial a_{i, 0}} &= \Phi,  \hspace{2em} \frac{\partial a_{\text{max}, 0}}{\partial a_{i, p}} = (a_{i, p}-a_{j, p})\frac{\phi}{\theta},  \hspace{2em} \frac{\partial a_{\text{max}, 0}}{\partial a_{i, r}} = a_{i, r}\frac{\phi}{\theta},  \\
\frac{\partial a_{\text{max}, q}}{\partial a_{i, 0}} &= (a_{i, q}-a_{j, q})\frac{\phi}{\theta}, \\
\frac{\partial a_{\text{max}, q}}{\partial a_{i, p}} &= -(a_{i, q}-a_{j, q}) (a_{i, 0}-a_{j,0}) (a_{i, p}-a_{j, p})\frac{\phi}{\theta^3}, \hspace{1em}p \neq q, \\
\frac{\partial a_{\text{max}, q}}{\partial a_{i, p}} &= -(a_{i, q}-a_{j, q}) (a_{i, 0}-a_{j,0}) (a_{i, p}-a_{j, p})\frac{\phi}{\theta^3}+\Phi,  \hspace{1em} p=q,\\
\frac{\partial a_{\text{max},q}}{\partial a_{i, r}} &= -a_{i, r}(a_{i, q}-a_{j, q}) (a_{i, 0}-a_{j, 0}) \frac{\phi}{\theta^3},\\
\frac{\partial a_{\text{max}, r}}{\partial a_{i, 0}} &= \frac{1}{2a_{\text{max}, r}}[2(a_{i,0}-a_{\text{max}, 0})\Phi+ \theta\phi \\ 
&+\frac{\phi}{\theta}(\sigma_{D_{i}}^2-\sigma_{D_{j}}^2-2\sum_{k=1}^m a_{\text{max}, k}(a_{i, k}-a_{j, k}))], \\
\frac{\partial a_{\text{max}, r}}{\partial a_{i, p}} &= \frac{a_{i, p}-a_{j, p}}{2a_{\text{max}, r}}[2(1-\Phi)+(a_{i, 0}+a_{j, 0}-2a_{\text{max}, 0})\frac{\phi}{\theta}+ \\ 
        &(a_{i,0}-a_{j,0})(\sigma_{D_j}^2-\sigma_{D_i}^2+2\sum_{k=1}^m a_{\text{max}, k}(a_{i, k}-a_{j, k}))\frac{\phi}{\theta^3}], \\
\frac{\partial a_{\text{max}, r}}{\partial a_{i, r}} &= \frac{a_{i, r}}{2a_{\text{max}, r}}[2\Phi+(a_{i, 0}+a_{j,0}-2a_{\text{max}, 0})\frac{\phi}{\theta}+ \\ 
        &(a_{i, 0}-a_{j, 0})(\sigma_{D_j}^2-\sigma_{D_i}^2+2\sum_{k=1}^m a_{\text{max}, k}(a_{i, k}-a_{j, k}))\frac{\phi}{\theta^3}].
\end{split}
\end{equation}
\subsubsection{ReLU and max-pooling layers}
For ReLU, the derivation of \textit{max} is applied directly since the \textit{max} is used independently among distributions.
For max-pooling, since the result is obtained by repeatedly applying the two input max operations, the gradients of the input distributions are obtained by iteratively applying the derivation of \textit{max} with the stored tightness probabilities.

\subsubsection{Partial derivative for batch normalization}
Since the reformulated batch normalization layer does not have distribution operations involved, the derivative follows the traditional approach.
Given the gradient of loss $L$, the gradients of sensitivities in distribution $D_{i}^{l}$ are
\begin{equation}
    \begin{split}
        \frac{\partial L}{\partial a_{i, k}^{l}} &= \frac{1}{m\sqrt{\sigma_{D_i^l}^2+\epsilon}}(m\frac{\partial L}{\partial \hat{a}_{i, k}^{l+1}}-\sum_{j=1}^m \frac{\partial L}{\partial \hat{a}_{j, k}^{l+1}} -\hat{a}_{i, k}^{l+1}\sum_{j=1}^{m}\frac{\partial L}{\partial \hat{a}_{j, k}^{l+1}}\hat{a}_{j, k}^{l+1}), \\
        \frac{\partial L}{\partial \gamma} &= \sum_{k=1}^{m}\frac{\partial L}{\partial a_{i, k}^{l+1}} \hat{a}_{i, k}^{l+1}, \hspace{3em}
        \frac{\partial L}{\partial \beta} = \sum_{k=1}^{m} \frac{\partial L}{\partial a_{i, k}^{l+1}}
    \end{split}
\end{equation}
where $\hat{a}_{j, k}^{l+1} = (a_{j, k}^{l+1}-\beta)/\gamma$ and $k\in\{1, m\}$.

\subsection{Training, Inference, and Complexity Analysis}
With back propagation as discussed above, the training can be easily carried out as follows.
The distributions are first extracted by ICA with a predefined extraction span. 
The extracted distributions then propagate through the constructed SCNN layers.
Before entering the evaluation module, the propagated distributions are unmixed to form a temporal feature map.
When the loss is obtained after evaluation with the proposed objective function, the error is propagated backward through the derived route.
The gradients of the canonical form distributions are calculated to act as the gradient outputs of the corresponding layers.
Then the weights with deterministic numbers are updated based on the obtained gradient outputs and the predefined learning rate.

The speedup of SCNN mainly comes from the fact that $N$ input images are modeled by a single parameterized canonical model of the same size. 
On the other hand, the computation complexity at each layer, including \textit{max}, \textit{sum} and assigning weights in forward propagation is increased by $\mathcal{O}(m)$. 
In addition, SCNN requires the extraction of the parameterized canonical model by ICA at the input, which incurs additional complexity overhead. 
Fortunately, with the fast ICA implementations available on GPUs, the execution time is negligible compared with the SCNN inference time \cite{ramalho2010efficient,kumara2016generalized}.  
As such, networks with SCNN backbone can achieve an inference speedup of approximately $N/m$. Such analysis is supported by our experiments later.

\subsection{Extension to Nonlinear Canonical Form}
Note that so far we have only discussed the linear parameterized canonical form obtained from ICA and its associated extension to various CNN layers.
It is also possible to obtain other nonlinear parameterized canonical form as suggested by \cite{singh2006statistical,cheng2009non} in a different context. 
We believe such an extension can be adopted for the proposed SCNN as well. 
For simplicity, we will not discuss it further in this paper but defer it as our future work.

\section{Video Object Detection: an Application}
\begin{table*}[!t]
\centering
\resizebox{\textwidth}{!}{%
\begin{tabular}{c|ccccc|ccccc|ccccc|ccccc}
\hline
Models & \multicolumn{5}{c|}{VGG11} & \multicolumn{5}{c|}{VGG16} & \multicolumn{5}{c|}{ResNet18} & \multicolumn{5}{c}{ResNet34} \\ \hline
$m$  & - & 8 & 10 & 12 & 14 & - & 8 & 10 & 12 & 14 & - & 8 & 10 & 12 & 14 & - & 8 & 10 & 12 & 14 \\
mAP  & \textbf{62.2} & \textbf{56.5} & 58.1 & 58.3 & 58.7 & \textbf{61.9} & \textbf{57.1} & 57.2 & 57.7 & 58.3 & \textbf{65.1} & \textbf{57.4} & 58.7 & 60.0 & 61.9 & \textbf{64} & \textbf{56.7} & 59.2 & 60.5 & 62.2 \\
FPS  & \textbf{101} & \textbf{181} & 150 & 130 & 116 & \textbf{65} & \textbf{116} & 98 & 84 & 72 & \textbf{247} & \textbf{363} & 340 & 304 & 276 & \textbf{161} & \textbf{246} & 225 & 210 & 188 \\ \hline
\end{tabular}
}
\caption{Accuracy/speed comparison between networks \textbf{with} SCNN backbone (configured with various basis dimension $m$ and fixed extraction span $N$=16) and \textbf{without} SCNN backbone (marked with $-$ in Row $m$).  
}
\label{table:exp_result}
\end{table*}
\begin{table*}[t]
\centering
\resizebox{16cm}{!}{%
\begin{tabular}{c|ccccccccccc}
\hline
Category & boat & building & car & drone & horseride & paraglider & person & riding & truck & wakeboard & whale  \\ \hline
w/o SCNN  & 90.0 & 91.3 & 60.7 & 50.1 & 40.5 & 60.1 & 58.6 & 80.4 & 30.5 & 28.8 & 90.4 \\
w/ SCNN   & 91.0 & 73.8 & \textbf{76.6} & 42.5 & 24.3 & 48.7 & 65.7 & \textbf{86.9} & 28.7 & 20.7 & 82.3 \\ \hline
\end{tabular}
}
\caption{Accuracy (mAP) across categories of VGG16 with SCNN backbone ($m=14$, $N=16$) and the one without it.}
\label{table:category_result}
\end{table*}

We believe SCNN can be a general and powerful backbone to any CNN networks and it processes parameterized statistical distributions rather than a deterministic values. 
Many CNN-based applications would benefit from such a representation. 
As a proof of point, we apply SCNN to the video object detection task to show its usefulness. 
Please note that, our initial implementation of SCNN (i.e., the statistical version of FC, CN, ReLU, Max-pooling and Batch normalization etc.) is far from perfection compared to those matured implementations in existing frameworks such as TensorFlow, Caffe, PyTorch. 
Because of that, our current implementation of SCNN to solve the video object detection is not yet optimized. 
Hence it is not our intention in this
paper to compete in either training performance or inference quality with the state-of-the-art video object detection techniques such as Faster R-CNN, YOLOv2/v3 etc \cite{ren2015faster,redmon2017yolo9000,yolov3}, although we have
shown the theoretic performance advantage. 
Instead, we want to use our implementation to show the great potential of SCNN for solving important computer vision problems and where it can potentially shine. 
In solving the video object detection problem, we proposed a few modifications to the commonly used object detection techniques in the context of SCNN.

We first replace a few commonly
used backbone CNN networks for
object detection with the proposed
SCNN, including VGG11, VGG16,
ResNet18 and ResNet34. We
then add a simple
evaluation module consisting
of conv-relu-conv-relu-conv
layers without padding. 
Because SCNN can effectively process multiple frames at the same time, a few changes need to be made when designing the detection layer and the objective function. 

For simplicity, we start with the case where there is only a single target object in videos and design a simplified detection layer based on YOLOv2 framework\cite{redmon2017yolo9000}.
In the detection layer of YOLOv2, predefined anchor boxes along with their confidence are predicted at each sub grid cell ($13\times 13$ total) to detect objects.
Such an approach is, however, not directly applicable to process
video snippets with a continuously moving object captured by a single canonical model. Therefore,
we propose a new detection layer with five predefined anchors ($a_{i, w}$, $a_{i, h}$ for $i\in\{0, 4\}$) at the center of the map (effectively treating the map as a single big cell).
The network predicts coordinates for the box ($t_x$, $t_y$, $t_w$, $t_h$) along with its confidence. These predictions in turn define
the predicted bounding box as follows:
\begin{equation}
\begin{split}
    b_x=\sigma(t_x), \hspace{1em} b_w&=\frac{a_{i, w}}{\beta}\log(1+\exp(\beta t_w)), \\
    b_y=\sigma(t_y), \hspace{1em} b_h&=\frac{a_{i, h}}{\beta}\log(1+\exp(\beta t_h))
\end{split}
\label{eq:yolo}
\end{equation}
where $\sigma$ denotes the sigmoid function and $\beta$ is for the formulation of Softplus.
Note that we use Softplus function to configure the width $b_w$ and height $b_h$ rather than direct exponential as was used in YOLOv2. This modication brings a more stable and smooth transformation on anchor size and fit well with our one big cell setting.

Since SCNN simultaneously handles multiple frames, the detection objective function should not only consider the precision on a single frames, but also account for the continuity of objects among adjacent frames. As such,  we propose a new objective function for SCNN, which is a combination of coordinates $\mathnormal{l}_2$ loss ($\mathcal{L}_{\text{coord}}$), confidence loss ($\mathcal{L}_{\text{conf}}$), polynomial fitting loss ($\mathcal{L}_{\text{fit}}$), and IOU loss ($\mathcal{L}_{\text{IOU}}$).

The IOU loss is first introduced in UnitBox\cite{yu2016unitbox}, which increases the accuracy by regressing the prediction box as a whole unit.
However, the curve of natural logarithm used in Unitbox has a steep slope, which is weak when the IOU gets high and needs fine-tuning.
Moreover, if we only use the IOU loss in the objective function, it would remain constant when the prediction box is out of the target area. This will not be helpful to improve the   convergence of training.
Intuitively, we would prefer an IOU loss that can compensate the coordinates loss to further increase the IOU.
Therefore, in this work, we propose to use a negative log sigmoid function of IOU.
Moreover, different from YOLOv2 where IOU is included in the confidence score, we use confidence loss $\mathcal{L}_{\text{conf}}$ to detect whether there is an object or not.

To further improve the accuracy, we observe that within the frames in the extraction span, the trajectories of object bounding box coordinates can
be approximated by a polynomial curve.
After predicting coordinates with Equation~\ref{eq:yolo}, we adopt the least-square polynomial fitting to obtain the corrected coordinates along with the fitting loss $\mathcal{L}_{\text{fit}}$.
The loss is then appended to the objective function as a penalty term.

In summary, given an initial bounding box prediction $z=(b_x, b_y, b_w, b_h, C_z)$, after fitting correction $\hat{z}$ and its corresponding ground truth $\tilde{z}$, the IOU between $z$ and $\tilde{z}$ marked as $X_{z, \tilde{z}}$, the objective function $\mathcal{L}(z,\tilde{z})$ is expressed as:
\begin{equation}
\begin{split}
\mathcal{L}(z,\tilde{z}) =& \lambda_{\text{coord}}\mathcal{L}_{\text{coord}} + \lambda_{\text{fit}}\mathcal{L}_{\text{fit}} + \lambda_{\text{conf}}\mathcal{L}_{\text{conf}} +  \lambda_{\text{IOU}}\mathcal{L}_{\text{IOU}} \\
=& \lambda_{\text{coord}}\sum_{\mathclap{i\in\{x, y, w, h\}}}(z_i-\tilde{z}_i)^2 + \lambda_{\text{fit}}\sum_{\mathclap{i\in\{x, y, w, h\}}}(z_i-\hat{z}_i)^2 \\
&+ \lambda_{\text{conf}}(\sum\mathbb{1}^{\text{obj}}(C_z-C_{\tilde{z}})^2 + \sum\mathbb{1}^{\text{noobj}}(C_z-C_{\tilde{z}})^2) \\
&-\lambda_{\text{IOU}}\ln(\frac{1}{1+\exp(-\alpha X(z, \tilde{z}))})
\end{split}
\end{equation}
where the subscripts $x, y, w, h$ represent the center coordinates, the width and height of the bounding box respectively;
$\lambda$ are the coefficients of loss terms;
$C$ is the objectness confidence score;
$\alpha$ in $\mathcal{L}_{\text{IOU}}$ is to adjust the IOU loss curve.

\subsection{Experiment Implementation Details}

We choose PyTorch as our evaluation platform to implement all models.
The experiments were run on 16 cores of Intel Xeon E5-2620 v4, 256G memory, and an NVIDIA GeForce GTX 1080 GPU. 
The dataset\cite{xu2018dac} is the latest video object detection dataset from the DAC 2018 system design contest. 
The dataset is challenging as videos are captured by drones in the air and the objects captured are small with a large variety in terms of its object classes, appearances, environment, and video qualities.

For accuracy, we use mean average precision (mAP) that calculates the ratio of IOU between predicted and ground truth bounding boxes larger than 0.5. 
Note that such a metric
is in fact not favorable to SCNN
because SCNN is able to process
and evaluate multiple image frames (video snippets) in one
pass, while the conventional
object detection is only able
to process one static image
at a time which has some
inherent accuracy advantage. 
Nonetheless, our comparison
will show that SCNN can achieve
a great speedup.

Overall the SCNN video object detection framework follows Figure~\ref{fig:scnn}.
The input image size is $224\times 224$ and $7\times 7$ temporal feature maps are obtained for evaluation.
The Stochastic Gradient Descent (SGD) solver is applied in SCNN training with an initial learning rate 0.001.
The momentum and weight decay are always set to 0.9 and 0.0005, respectively.

We then implement VGG \cite{simonyan2014very} and ResNet both with and without SCNN backbone for accuracy and speed comparisons. VGG is known for its simple sequential network which only uses $3\times 3$ stacked convolutional layers for feature extraction. 
ResNet is characterized by its network-in-network structure which leads to effective extremely deep network. 
For implementation with SCNN, all the layers for feature extraction in these networks are redesigned according to the previous discussion. 
For classifiers in VGG and ResNet, the original fully connected layers are replaced with the evaluator discussed previously. 
The number of kernels in the evaluation module is updated according to the output of the corresponding network.
All networks are trained from scratch with the same optimizer setting.

\subsection{Results}
The video object detection accuracy and speed for the networks with SCNN backbone using different basis dimension $m$ and the same extraction span ($N=16$), along with their counterparts without SCNN backbone are shown in Table~\ref{table:exp_result}.  
From the table we can see that networks with SCNN backbone can achieve higher inference speed with slight accuracy degradation.  
For example, when m=8, VGG16 with SCNN backbone can achieve a speedup of $178\%$ with a $4.8\%$ drop in mAP compared with the one without it. This fully demonstrates the efficiency of SCNN. 
Also, with larger basis dimension, networks with SCNN backbone tend to achieve better accuracy at the cost of lower inference speed.  

To further illustrate the performance of SCNN, we take VGG16 as an example and compare the mAP of VGG16 with and without SCNN backbone across multiple categories in the dataset. The results are shown in Table~\ref{table:category_result}.

Although SCNN can achieve reasonable accuracy as CNN with higher FPS, we see from the Table~\ref{table:exp_result} that SCNN has lower mAP than CNN. 
By looking into the details in Table~\ref{table:category_result}, we find that SCNN in fact outperforms CNN for object categories that are relatively smooth across frames such as car and riding. 
This is because SCNN can mitigate object occlusion and lens flare effects with its implicit modeling of temporal correlations via ICA. 
In contrast, for objects such as building, paraglider or horseride that are either too large or too small, the errors due to ICA as shown in Fig~\ref{fig:error_hist} start to have a negative impact. 
Rather than to use a linear parameterized form as obtained by ICA, a direction for future improvement will be to use the nonlinear parameterized distribution that can model large-scale spatial correlation more explicitly. 
Another possible direction is to explore the SCNN specific network architecture rather than piggyback on existing CNN architecture.

\section{Conclusion and Discussion}
\label{sec:con}
In this paper we proposed a novel statistical convolutional neural network (SCNN), which operates on distributions in parameterized canonical model.  
Through a video object detection example, we show that SCNN as an extension to any existing CNNs can process multiple correlated images effectively, achieving great speedup over existing approaches.

The performance of SCNN can be further enhanced by utilizing the correlations not only between adjacent frames in a video snippet but also among channels of the same frame. This shall provide further speedup. Such a change may require, however, redesigning the CNN network topology because the input dimension is now different, a future research direction worthy to explore. It will  be also interesting to see how SCNN can be used in other applications such as uncertainty-aware image classification or segmentation.

\fontsize{9.5pt}{10.5pt} \selectfont  
\bibliography{bib.bib}
\bibliographystyle{aaai}
\end{document}